\documentclass[onecolumn,10pt]{article}
\usepackage[top=1.25in, bottom=1.25in, left=1.25in, right=1.25in]{geometry}
\usepackage{amsmath,amsfonts,amscd,amssymb}
\usepackage{graphicx}
\usepackage{overpic}
\usepackage{cancel}
\usepackage{rotating}
\usepackage{url}
\usepackage{caption}
\usepackage{color}
\usepackage{rotating}
\usepackage{multirow}
\usepackage{wrapfig}
\usepackage{mathtools}
\usepackage{subeqnarray}
\usepackage{setspace}
\usepackage{longtable,tabularx}
\usepackage{geometry}
\usepackage{multicol}
\usepackage{palatino} 
\usepackage[normalem]{ulem}

\usepackage{float}
\graphicspath{{figures/}}

\usepackage[bottom,flushmargin,hang,multiple]{footmisc}

\newcommand\blfootnote[1]{%
  \begingroup
  \renewcommand\thefootnote{}\footnote{#1}%
  \addtocounter{footnote}{-1}%
  \endgroup
}
\usepackage{hyperref}
\renewenvironment{abstract}
 {\par\noindent\rule{\linewidth}{.25pt}\par\noindent\textbf{\abstractname}\par\noindent \ignorespaces}
 {\par\noindent\medskip\rule{\linewidth}{.25pt}}
 
\newcommand{\be}{\mathbf{e}}

\newcommand{\bu}{\mathbf{u}}

\newcommand{\bx}{\mathbf{x}}
\newcommand{\by}{\mathbf{y}}
\newcommand{\bn}{\mathbf{n}}
\newcommand{\bz}{\mathbf{z}}

\newcommand{\bX}{\mathbf{X}}

\newcommand{\bbR}{\mathbb{R}}
\newcommand{\bbD}{\mathbb{D}}

\newcommand{\bbE}{\mathbb{E}}
\newcommand{\bI}{\mathbf{I}}

\newcommand{\cO}{\mathcal{O}}

\newcommand{\cD}{\mathcal{D}}

\usepackage{titlesec}
\titleformat*{\section}{\bfseries}
\titleformat*{\subsection}{\itshape}
\titleformat*{\subsubsection}{\itshape}
\titleformat*{\paragraph}{\large\bfseries}
\titleformat*{\subparagraph}{\large\bfseries}

\title{\Large{\vspace{-.65in}{Output-weighted and relative entropy loss functions for deep learning precursors of extreme events}}}
\author{\normalsize{Samuel H. Rudy$^*$ and Themistoklis Sapsis} \\ 
\footnotesize{Department of Mechanical Engineering, Massachusetts Institute of Technology, Cambridge, MA 02139} \vspace{-1 mm} \\
}
\date{}

\begin{document}

\maketitle

\blfootnote{$^*$ Corresponding author (shrudy@mit.edu)}
\vspace{-.25in}

\begin{abstract}
Many scientific and engineering problems require accurate models of dynamical systems with rare and extreme events. 
Such problems present a challenging task for data-driven modelling, with many naive machine learning methods failing to predict or accurately quantify such events. 
One cause for this difficulty is that systems with extreme events, by definition, yield imbalanced datasets and that standard loss functions easily ignore rare events. 
That is, metrics for goodness of fit used to train models are not designed to ensure accuracy on rare events. 
This work seeks to improve the performance of regression models for extreme events by considering loss functions designed to highlight outliers.
We propose a novel loss function, the adjusted output weighted loss, and extend the applicability of relative entropy based loss functions to systems with low dimensional output. 
The proposed functions are tested using several cases of dynamical systems exhibiting extreme events and shown to significantly improve accuracy in predictions of extreme events. \\

\vspace{-0.05in}
\end{abstract}

\section{Introduction} \label{sec:intro}

Accurate prediction and quantification of extreme events are critical tasks in many areas of science and engineering. 
Specific cases include the study of extreme weather patterns \cite{easterling2000observed}, turbulence \cite{yeung2015extreme}, macroeconomic fluctuations \cite{longin1996asymptotic}, rouge waves \cite{dysthe2008oceanic}, and many others \cite{sapsis2021statistics, farazmand2019extreme}. 
Recent research has developed tools for experimental design aimed towards uncertainty quantification in tail regions of systems with extreme events \cite{mohamad2018sequential, sapsis2020output, blanchard2021bayesian}, prediction in the sense of classification of upcoming events in turbulent fluid flows \cite{farazmand2017variational,blonigan2019extreme,guth2019machine}, and regression problems for systems exhibiting extreme events \cite{doan2021short, qi2020using}.  

These recent works fall into the broader category of data-driven approaches to dynamical systems and fluid dynamics \cite{brenner2019perspective, brunton2020machine}. Uses for such methods are motivated by cases in which physics based models fail due to intractable complexity, computational requirements, or insufficient measurements.  In these cases machine learning and in particular deep learning offer a potential tool for improved predictive modelling. Successful applications of deep learning to problems in dynamical systems include flow reconstruction \cite{milano2002neural}, physics informed neural networks \cite{raissi2019physics}, closure models \cite{duraisamy2019turbulence,gupta2020neural}, sub-grid scale models \cite{brenowitz2018prognostic, rasp2018deep}, climate modelling \cite{irrgang2021towards}, operator inference \cite{lu2021learning}, and embedding and lifting transformations \cite{qian2020lift, lusch2018deep, champion2019data}.

Several recent works have explored the use of specific loss functions for training prediction models in the context of extreme events. 
Guth and Sapsis \cite{guth2019machine} develop the maximum adjusted area under the precision recall curve and show that it is effective in predicting extreme events in systems including the Kolmogorov flow and Majda-McLaughlin-Tabak model \cite{majda1997one}.
However, their proposed metric is not differentiable and not well approximated by small samples. 
Implementation for high dimensional models such as neural networks would therefore be challenging.
Doan et. al. \cite{doan2021short} use a physics informed loss function improve the accuracy of echo-state networks for forecasting a Galerkin model of a turbulent flow with intermittent quasi-laminar states. 
While effective, this approach is constrained to problems where there is a known dynamic model for the quantity being predicted. 
Authors of \cite{qi2020using} use a relative entropy based loss function to forecast the truncated Korteweg–de Vries equation, a simplified model of turbulent surface waves. 
This is shown to significantly improve performance, but requires a high dimensional target quantity. 
More recently, the use of various model architectures for predicting extreme events has been studied in \cite{asch2021model}.

In this work we seek to develop more broadly applicable loss functions and evaluate their performance on several challenging test problems.
While the loss functions proposed in this work may be applied to arbitrary regression models, the included examples problems both employ neural networks. 
We assume the reader is familiar with common deep learning techniques including recurrent neural networks, stochastic optimization, and early stopping. 
The unfamiliar reader may find an excellent and free online reference in \cite{goodfellow2016deep}. 
In particular, we make use of long-short-term-memory networks \cite{hochreiter1997long} for each of the test cases used in this work.
The results could almost certainly be improved on via more carefully thought out network structures, training, and other user decisions \cite{asch2021model}. 
However, such considerations are not the focus of this work which focuses solely on the effect of loss functions. 

The paper is organized as follows; 
In Sec. \ref{sec:methods} we outline the motivation for extreme event specific loss functions and develop methods including output weighted variations of the mean square error and a relative entropy method based on work in \cite{qi2020using}. 
We also discuss error metrics for evaluating the proposed loss functions.
In Sec. \ref{sec:results} we present results of the proposed loss functions applied to two test cases; Kolmogorov flow at Reynolds number $Re=40$ and the flow around a square cylinder at $Re=5000$. 
A discussion of the results, limitations, and outlook is presented in Sec. \ref{sec:discussion}.
\section{Methods} \label{sec:methods}

In this section we outline the proposed loss functions used for training neural networks to predict and quantify extreme events.  These include two weighted variations of the mean square error as well as a relative entropy, also known as KL-divergence, based loss.  We also describe methods for approximating the density function of the target variable $y$ and the metrics we use to measure accuracy of the trained networks.

\subsection{Problem statement}\label{subsec:problem_statement}

Let $(X, \Sigma_x, \mu)$ be a probability space with $X \subseteq \bbR^n$ and $\mu$ absolutely continuous with probability density function $p_x$. 
For some unknown function $f$, we have a dataset $\cD = (\bX,\by)$ where $\bX = \{\bx_1, \hdots , \bx_m\}$, $\bx_i \sim p_x$, $y = \{y_1, \hdots, y_m\}$ with $y_i = f(\bx_i)$, perhaps perturbed by measurement noise. 
We will make use of the fact that since $p(\cD) = \prod_{i} p_x(\bx_i)$ we have $\bbE_\cD[\frac{1}{m} \sum g(\bx_i)] = \bbE_\bx [g(\bx)]$ for any $g$ such that the expectation is finite.  That is, empirical averages over $\cD$ are unbiased approximations of integrals over $p_x$.

We are interested in parametric models $\hat{f}$ approximating $f$ which accurately predict and quantify outlier values of $y$ and which accurately capture the induced probability density function through a measure transformation: $p_y(y) = d/dy\,\mu (f^{-1} (-\infty, y))$.
To this end, both the structure of the model and training parameters have important effects. 
We seek to develop objective functions tailored for extreme event prediction that accurately capture extreme events, are differentiable, and may be approximated from finite datasets. 

Throughout the remainder of the paper, we will use the terms true and false positive and negative to describe various results from a continuous regression model. 
In this context, we loosely define a true positive to be a prediction $\hat{f}(\bx)$ such that $p_y( f (\bx) )\sim p_y( \hat{f} (\bx) )\ll 1$.  
That is, when both the true and predicted values of $y$ are rare. 
Likewise, a true negative is when $p_y( f (\bx) ) \sim p_y ( \hat{f} (\bx) )\sim \cO(1)$, a false positive is when $p_y (f (\bx)) \sim \cO(1)$ and $p_y ( \hat{f} (\bx)) \ll 1$, and a false negative is when $p_y ( f (\bx) ) \ll 1$ and $p_y ( \hat{f} (\bx)) \sim \mathcal{O}(1)$. 
In some places, these definitions will be made rigorous by applying thresholds to $p_y(f (\bx))$ and $p_y(\hat{f} (\bx))$.

\subsection{Output-weighted variations of mean square error}\label{subsec:output_weighted_mse}

The mean square error is the most common loss function used for training regression models with real valued outputs.  It is given by,
\begin{equation}
L_{MSE}(\hat{f}) =\bbE_\bx\left[ e_{\hat{f}}(\bx)^2 \right] = \int_{X} e_{\hat{f}}(\bx)^2 p_x(\bx) \, d\bx = \bbE_\cD \left[ \frac{1}{m} \sum_\cD e_{\hat{f}}(\bx_i)^2 \right],
\label{eq:MSE}
\end{equation}
where $e_{\hat{f}} = (f(\bx) - \hat{f}(\bx))$.  Squaring the error makes $L_{MSE}$ more sensitive to true outliers than the mean absolute error, but if outlier values make up a small fraction of the total dataset then $L_{MSE}$ may still be small while missing large $y$.  Moreover, if some rare values of $y$ are not separated from the core of $p_y$ by substantial distance then they may be missed with little added error.  To better see this, consider the case where $\mu$ admits a disintegration over the induced measure on $y$. 
Then we can express $L_{MSE}$ as an integral over $Y$ of the regular conditional expectation of the square error. That is,
\begin{equation}
L_{MSE}(\hat{f}) = \bbE_y\left[ \bbE_\bx\left[ \left. e_{\hat{f}}(\bx)^2 \right| f(\bx) = y \right]\right]  = \int_{Y} p_y(y) \, \bbE_\bx\left[\left. e_{\hat{f}}(\bx)^2 \right| f(\bx) = y \right] \, dy,
\label{eq:MSE_Y}
\end{equation}
where the conditional expectation is defined using the disintegration of $p_x$ over $p_y$ \cite{chang1997conditioning}.  For rare events, the value of $p_y(y)$ is small, allowing large error in the prediction of such events without significantly affecting $L_{MSE}$.

The insensitivity of $L_{MSE}$ to rare events may be mitigated via introducing a weighting function.  Specifically, consider the case where the square error is weighted according to $p_y(f(\bx))^{-1}$.  The resulting function is given by,
\begin{equation}
\begin{aligned}
L_{OW}(\hat{f}) &= \bbE_\bx\left[ p_y(f(\bx))^{-1}e_{\hat{f}}(\bx)^2 \right]\\
&= \int_{Y} p_y(y) \, \bbE_\bx\left[\left. p_y(f(\bx))^{-1}e_{\hat{f}}(\bx)^2 \right| f(\bx) = y \right] \, dy \\
&= \int_{Y} \, \bbE_\bx\left[\left. e_{\hat{f}}(\bx)^2 \right| f(\bx) = y \right] \, dy .
\end{aligned}
\label{eq:OW}
\end{equation}
Note that the expression $p_y(f(\bx)) |_{f(\bx)=y}$ is simply $p_y(y)$ and therefore cancels the $p_y(y)$ term outside the conditional expectation. 
We call the expression given by Eq. \eqref{eq:OW} the output-weighted loss, $L_{OW}$ since the square error is weighed by the inverse of the likelihood of the true output $f(\bx)$.  
Expressions with similar form have been used for sequential sampling strategies for rare events \cite{sapsis2020output,blanchard2021bayesian}. 
However, these works were focused on experimental design and used least squares or Gaussian process regression to model $f$. 
Equation \eqref{eq:OW} is also related to oversampling techniques commonly used for classification problems with imbalanced data \cite{he2009learning}. 

As a cost function, Eq. \eqref{eq:OW} has some potential drawbacks. 
In particular, the weight given to each error is proportional only to the inverse likelihood of the true value, $p_y ( f(\bx))$, and is independent from $p_y ( \hat{f}(\bx) )$. 
Thus, error accumulated on false negatives is penalized far more than that made on false positives. 
As we will show in Sec. \ref{sec:results}, minimizing Eq. \eqref{eq:OW} often yields models that over-predict rare events. 

\begin{figure}[ht!]
\centering
\includegraphics[width=\textwidth]{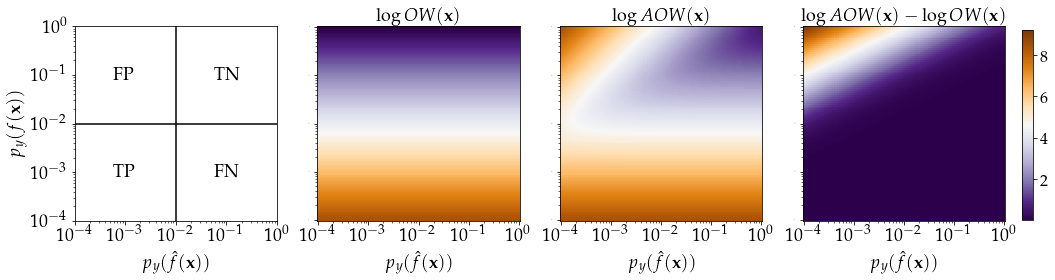}
\caption{Visualization of weights used by $L_{OW}$ and $L_{AOW}$.  Left most figure shows hypothetical boundaries for true/false positive/negative regions.  Pseudocolor plots show output weight, adjusted output weight, and difference.  Disagreement is primarily within the false positive region.}
\label{fig:weights_pcolor}
\end{figure}

The problem of false positives may be addressed with the inclusion of a second term weighing the square error.
The weight function $1 / p_y(f(\bx))$ in Eq. \eqref{eq:OW} amplifies any error realized on examples with rare $y$. 
This included true positive and false negative predictions. 
To distinguish between true negative and false positive predictions, we require a weight function that depends on $p_y(\hat{f}(\bx))$.
The ratio $p_y(f(\bx))/p_y(\hat{f}(\bx))$ is large only in the case where the likelihood of the predicted output is lower than that of the true output. 
Thus, the expression given by, 
\begin{equation}
\begin{aligned}
L_{FP}(\hat{f}) &= \bbE_\bx\left[ \frac{p_y(f(\bx))}{p_y(\hat{f}(\bx))}e_{\hat{f}}(\bx)^2 \right],
\end{aligned}
\label{eq:FP_error}
\end{equation}
is a measure of the error made on false positive predictions.  Summing Eq. \eqref{eq:OW} with Eq. \eqref{eq:FP_error} gives,
\begin{equation}
\begin{aligned}
L_{AOW}(\hat{f}) &= L_{OW}(\hat{f}) + L_{FP}(\hat{f}) \\
&= \bbE_\bx\left[ \left( \dfrac{1}{p_y(f(\bx))} +\dfrac{p_y(f(\bx))}{p_y(\hat{f}(\bx))} \right) e_{\hat{f}}(\bx)^2 \right] 
\end{aligned}
\label{eq:AOW}
\end{equation}
which we will call the \textit{adjusted output weighted loss}, or $L_{AOW}$.
For true positives, when both $p_y ( f (\bx) )$ and $p_y ( \hat{f}(\bx) )$ are small, the second term in the parentheses is $\cO(1) \ll p_y ( f(\bx))^{-1}$.  
The integrand therefore agrees with that of Eq. \eqref{eq:OW}.  For true negatives and false negatives, the integrand is also of similar magnitude to that in Eq. \eqref{eq:OW}.  
However, for false positives, the second term in the parenthesis is large.  
The expression therefore penalizes errors made as false positive predictions to a greater degree than Eq. \eqref{eq:OW}. 

The added penalization of false positives comes at the cost of increased complexity.  
Equation \eqref{eq:OW} is a weighted least squares problem and does not add any computational complexity to Eq. \eqref{eq:MSE} beyond pre-computing $\{p_y(y_i)\}_{i=1}^m$. 
In particular, Eq. \eqref{eq:OW} has closed form solution for linear problems. 
This is not the case for Eq. \eqref{eq:AOW}, where iterative optimization is required.
For $\hat{f}$ depending nonlinearly on parameters, such as neural networks, the difference is less important, as iterative methods must be used in either case.

Differences between the weighting functions in the integrands of Eq. \eqref{eq:OW} and \eqref{eq:AOW} are visualized in Fig. \ref{fig:weights_pcolor}.  We plot values of the each term using hypothetical values of $p_y( f (\bx))$ and $p_y ( \hat{f} (\bx))$.  The left most plot illustrates a rough partitioning of the domain into true/false positives and negatives.
Note that many expressions of the likelihoods of the true and predicted $y$ could be constructed to find novel loss functions.  The authors suspect that this would be a fruitful research direction, but it is not within the scope of this work.


\subsection{Approximating $p_y$}\label{subsec:approx_py}

Computation of the two weighted variations of the mean square error in the previous section requires evaluating  $p_y$ at points $y_i$ in the training set.  For Eq. \eqref{eq:OW}, this may be done offline using any off-the-shelf density estimation technique.  
Evaluating \eqref{eq:AOW} further requires $p_y\circ\hat{f}$, which changes during training and must be evaluated rapidly.  
Gradient based optimization also requires evaluating $p_y' \circ \hat{f}$. 
Thus, minimizing Eq. \eqref{eq:AOW} requires a low-computational-cost differentiable approximation of $p_y$.
This rules out non-parametric density estimates such as kernel density estimation (due to slowness), histograms (due to lack of differentiability), and k-nearest-neighbors (slow and not differentiable). 
We instead use a combined method of nonparametric estimation of $log(p_y(y_i))$ which we then fit using a Gaussian process with few collocation points, allowing for rapid evaluation of $p_y$ and its derivative.

Estimates of $p_y(y_i)$ are initially made using a histogram of $\{y_i\}_{i=1}^m$ with $n_b=100$ evenly sized bins.   Defining $B_i = [b_i, b_{i+1})$ for $i=1,\hdots, 99$ and $B_{100} = [b_{100}, b_{101}]$ where $b_i$ are evenly spaced values between $min(y_i)$ and $max(y_i)$ we have for $y \in B_i$,
\begin{equation}
\log \hat{p}(c_i) \approx \log \left(  \frac{\left| \left\{ j : y_j \in B_i \right\} \right|}{m l }\right)
\label{eq:hist_estimate}
\end{equation}
where $l = b_{i+1} - b_i$ and $c_i = (b_i + b_{i+1})/2$.

Following estimation of the log densities via histogram, the values at the center of each bin are fit to a Gaussian process \cite{rasmussen2003gaussian}. 
We use the Matern-5\/2 kernel with additional white heteroscedastic noise. 
The mean function of the Gaussian process is set to be the log of machine epsilon, enforcing that the approximation goes to zero away from the sampled data. 
That is,
\begin{equation}
\log \hat{p}_y(y) \sim \mathcal{GP} \left( \mu, k(y,y') + \sigma_w^2(y) \delta_{y,y'} \right),
\end{equation}
where $\mu = \log(10^{-16})$, $\sigma_w(y)$ is the white noise term and $\delta_{y,y'}$ is the Kronecker delta function.
Fitting the logarithm of the probability rather than the probability helps to get accurate estimates in the tails of $p_y$.  
We obtain estimates of the density at sample points $y$ using the conditional mean of the Gaussian process;
\begin{equation}
\begin{aligned}
\hat{p}_y(y) &= \exp\left(k(y,\mathbf{c}) \left( k(\mathbf{c},\mathbf{c}) + \sigma_w^2(\mathbf{c})\bI \right)^{-1} \log\hat{p}_y(\mathbf{c}) + \mu \right) \\
& =  \exp\left(k(y,\mathbf{c}) \boldsymbol{\alpha} + \log(10^{-16}) \right).
\end{aligned}
\end{equation}
where $c_i = (b_i+b_{i+1})/2$ and vector $\boldsymbol{\alpha}$ is pre-computed and stored.  The expected gradient of the exponent is given by simply differentiating the kernel \cite{mchutchon2013differentiating}.
This allows for queries $\hat{p}_y (y)$ and it's gradient with the simple computation of $k(y,\mathbf{c})$ and $k'(y,\mathbf{c})$.
In practice we found training to be more stable when a floor was set for the value of $p_y(y)$.  
Examples in this work all used an effective density equal to $\hat{p}_y(y)+10^{-5}$ which has negligible effect of most events.

We note that the same process could be easily implemented for $y$ with dimension greater than one but still low.  Initial estimates of $p_y$ at training points of the Gaussian process could be taken with any standard non-parametric density estimation \cite{wasserman2006all} and subsequently fit to a GP.
However, for higher dimensional $y$, both density approximation and Gaussian process interpolation become non-trivial.
It is possible that in these cases the density $p_y$ could be substituted for that of a relevant observable $g(y)$, but such work is beyond the scope of the present manuscript.

\subsection{A relative entropy based loss function}\label{subsec:entropy_loss}

The use of relative entropy as a loss function for neural networks was explored in \cite{qi2020using}.  
Qi and Majda used the relative entropy (i.e. KL-divergence) between truth and prediction, after applying the soft-max function. 
Specifically, for $y \in \bbR^s$, \cite{qi2020using} uses a loss function defined for a single datapoint by,
\begin{equation}
\begin{aligned}
L_{QM}(\bx) &= KL \left( \sigma (f(\bx)) \left\| \sigma (\hat{f}(\bx)) \right. \right) + \alpha KL \left( \sigma (-f(\bx)) \left\| \sigma (-\hat{f}(\bx)) \right. \right) \\
&=\sum \left( \sigma(f (\bx)) \log \left( \frac{\sigma(f (\bx))}{\sigma(\hat{f} (\bx))} \right) + \alpha \sigma(-f (\bx)) \log \left( \frac{\sigma(-f (\bx))}{\sigma(-\hat{f} (\bx))} \right) \right)
\end{aligned}
\label{eq:L_QM}
\end{equation}
where $\log$ is taken elementwise, the sum is over $s$ dimensions of the vector enclosed in the parenthesis, and the soft-max function, $\sigma$, is defined by,
\begin{equation}
    \sigma(y) = \frac{exp(y)}{\sum exp(y)}
    \label{eq:softmax}
\end{equation}

The use of $\sigma$ weights outputs by the exponent of their magnitude, thus ensuring the loss focuses on accurate learning of large magnitude features.
This approach is extremely effective in \cite{qi2020using}, where $y$ the solution to a PDE and thus high dimensional.  However, it may not be applied directly in the case of scalar output $y$.  This is because for any $y \in \bbR^1$, $\sigma(y) = 1$. 
Moreover, Eq. \eqref{eq:L_QM} is only able to weight extreme values within one output sample, rather than comparing multiple $y$. 
However, it is possible to derive similar loss functions where a soft-max like operator is applied to multiple samples, rather than the indices within a single sample.

Let us assume that $f$ and $\hat{f}$ are such that $\bbE_x [ exp(f(\bx)) ]$ and $\bbE_x [ exp( \hat{f}(\bx))]$ are finite and non-zero.  Note that this is a mild assumption that holds on a set containing $L^\infty(X)$.  Then we can define an operator $G(f)$ by,
\begin{equation}
\begin{aligned}
G(f)(\bx) &= e^{f(\bx)} p_x(\bx) \left( \int_X e^{f(\bx)} p_x(\bx) \, d\bx  \right)^{-1} \\
&=e^{f(\bx)} p_x(\bx) \, \bbE_x \left[ e^{f(\bx)}  \right]^{-1}.
\end{aligned}
\label{eq:exp_tilt}
\end{equation}
This operator acts as a continuous analog of the soft-max function. 
Functions in the range of $G$ are probability density functions on $X$ where the value of $G(f)(\bx)$ is proportional to the sample density $p_x(\bx)$ and the exponent of $f(\bx)$. 
$G(f)$ therefore has mass concentrated on those values of $\bx$ whose likelihood under $p_x$ is not vanishing and whose image under $f$ is large, or in other words, extreme. 
Note that in the case where $f$ is linear the operator defined in Eq \eqref{eq:exp_tilt} is known as exponential tilting \cite{efron1981nonparametric}, which has previously been used in importance sampling \cite{siegmund1976importance}. 
We define the relative entropy loss as, 
\begin{equation}
L_{RE}(\hat{f}) = KL\left(\left. G(f) \right\| G(\hat{f}) \right).
\label{eq:RE}
\end{equation}
Note that compared to Eq. \eqref{eq:L_QM} used in \cite{qi2020using}, the normalization of $G(f)$ used in Eq. \eqref{eq:RE} is taken across the input space $X$ rather than dimensions of $y$. 
This allows for exponential weighting of outputs by their magnitude even in the case of scalar $y$. 
We are interested in minimizing Eq. \eqref{eq:RE} with respect to $\hat{f}$.  Expanding Eq. \eqref{eq:RE} and ignoring terms whose value does not depend on $\hat{f}$ we find,
\begin{equation}
\begin{aligned}
L_{RE}(\hat{f})  &= \bbE_\bx \left[ \dfrac{e^{f(\bx)} }{\bbE_\bx \left[ e^{f(\bx)}  \right]}   \log \left(  \dfrac{e^{f(\bx)} p_x(\bx) \, \bbE_\bx \left[ e^{\hat{f}(\bx)}  \right] }{e^{\hat{f}(\bx)} p_x(\bx)\, \bbE_\bx \left[ e^{f(\bx)}  \right]} \right) \right] \\
&\propto  \bbE_\bx \left[e^{f(\bx)}   \log \left(  e^{f(\bx) - \hat{f}(\bx)} \, \bbE_\bx \left[ e^{\hat{f}(\bx)}  \right] \right) \right] \\
&=  \bbE_\bx \left[e^{f(\bx)} (f(\bx) - \hat{f}(\bx)) + e^{f(\bx)}  \log \left( \bbE_\bx \left[ e^{\hat{f}(\bx)}  \right] \right) \right] .
\end{aligned}
\end{equation}
Individual expectations in the above expression may be estimated with sums over the dataset.  However, the term inside the log is problematic.   By Jensen's inequality,
\begin{equation}
\log \left( \bbE_\bx \left[ e^{\hat{f}(\bx)} \right] \right) = \log \left( \bbE_\cD \left[ \frac{1}{m} \sum_{i=1}^m e^{\hat{f}(\bx_i)} \right] \right) \geq \bbE_\cD \left[ \log \left(  \frac{1}{m} \sum_{i=1}^m e^{\hat{f}(\bx_i)} \right) \right]
\label{eq:log_exp_exp_jensen}
\end{equation}
where the second and third expectations are over the random samples $\cD$.  Thus,  the expected log of the empirical average of $exp(\hat{f}(\bx))$ is an underestimate.  We therefore find an upper bound for Eq. \eqref{eq:log_exp_exp_jensen} that may be accurately approximated and minimized.  Note that for any $\alpha$, $\log$ is bounded above by its first order Taylor expansion about $\alpha$.  Therefore,
\begin{equation}
\begin{aligned}
\log \left( \bbE_\bx \left[ e^{\hat{f}(\bx)} \right] \right) &= \log \left( \alpha + \left( \bbE_\bx \left[ e^{\hat{f}(\bx)} \right]- \alpha\right) \right) \\
&\leq \log(\alpha) +\dfrac{ \bbE_\bx \left[ e^{\hat{f}(\bx)} \right]- \alpha}{\alpha}.
\end{aligned}
\end{equation}
The error in the Taylor expansion and thus tightness of the bound is on the order of $(\bbE_\bx \exp(\hat{f}(\bx)) - \alpha)/\alpha^2$.   We therefore want to pick some $\alpha$ as close to $\bbE_\bx \exp(\hat{f}(\bx))$ as possible.  Consider $\alpha = \bbE_\bx \exp(f(\bx))$ which for $f\approx \hat{f}$ we assume will be close.  In this case the upper bound for $L_{RE}$ simplifies dramatically to,
\begin{equation}
\begin{aligned}
L_{RE}(\hat{f})  &\leq  \bbE_\bx \left[e^{f(\bx)} (f(\bx) - \hat{f}(\bx)) + e^{f(\bx)}  \left(  \log\left(  \bbE_\bx \left[ e^{f(\bx)} \right] \right) +\dfrac{ \bbE_\bx \left[ e^{\hat{f}(\bx)} \right]- \bbE_\bx \left[ e^{f(\bx)} \right]}{\bbE_\bx \left[ e^{f(\bx)} \right]}  \right) \right]  \\
&= \bbE_\bx \left[e^{\hat{f}(\bx)} - e^{f(\bx)} \hat{f}(\bx) \right],
\end{aligned}
\end{equation}
where we have ignored terms that do not depend on $\hat{f}$.  In particular, the log expectation term, $\log(\alpha)$, has been removed since it does not depend on parameters.  The remaining terms are easily approximated from dataset $\cD$ by,
\begin{equation}
\begin{aligned}
L_{RE}(\hat{f})  = \bbE_\bx \left[e^{\hat{f}(\bx)} - e^{f(\bx)} \hat{f}(\bx) \right] = \bbE_\cD\left[ \frac{1}{m} \sum_\cD \left(e^{\hat{f}(\bx_i)} - e^{f(\bx_i)} \hat{f}(\bx_i) \right) \right].
\end{aligned}
\label{eq:RE_D}
\end{equation}
Following \cite{qi2020using} we note that the relative entropy loss only focuses the error on large positive values of $f(\bx)$ and introduce the generalization,
\begin{equation}
L_{RE, \lambda}(\hat{f}) = L_{RE}(\hat{f}) + \lambda L_{RE}^{(-)}(\hat{f}) ,
\label{eq:SRE}
\end{equation}
where $ L_{RE}^{(-)}(\hat{f})$ is defined by replacing $f$ and $\hat{f}$ in Eq. \eqref{eq:RE_D} with $-f$ and $-\hat{f}$ respectively.  The value of $\lambda$ is a tuning parameter that can be set according to the skew of the dataset.  Data considered in this work contains extremes that skew positive. We therefore set $\lambda=0.1$, so that the loss function focuses on predicting positive outliers.

\subsection{Performance measures for regression with extreme events}\label{subsec:error_measures}

Goodness of fit in the context of regression for extreme events is non-trivial to define.  In Sec. \ref{subsec:problem_statement} we outlined three criteria for a ``good'' predictor.  That is, we seek models that accurately predict extreme events (in the sense that they may be used as a classifier), quantify extreme events, and yield accurate densities in the tails of $p_y$.  In this section we present metrics for quantifying each of the three criteria and discuss potential shortcomings of our approach towards error analysis.

\subsubsection{Accurate quantification of tails of $p_y(y)$}
Finally, we seek models such that the push-forward density under the learned model is similar to the true density $p_y$.  We are particularly interested in loss functions yielding densities that match the tails of the true density.  Following \cite{sapsis2020output,blanchard2021bayesian} use the difference between logarithms of the two density functions over the intersection of their support.  This is normalized by the size of the intersection to penalize distributions that only intersect on a small domain.  The metric is given by;
\begin{equation}
\mathbb{D}(p_y, \hat{p}_y) = \frac{1}{|\Omega(p_y,\hat{p}_y)|^2}\int_{\Omega(p_y,\hat{p}_y)}\left| \log(p_y(y)) - \log(\hat{p}_y(y)) \right|\, dy,
\label{eq:log_diff_pdf}
\end{equation}
where
\begin{equation}
\Omega(p_y,\hat{p}_y) \approx supp(p_y) \cap supp(\hat{p}_y).
\end{equation}
and $\hat{p}_y$ is the density under the learned model.  Since $p_y$ and $\hat{p}_y$ are approximated from data, their support is unknown and behavior in low density regions extremely challenging to quantify.  Unfortunately, $\mathbb{D}$ depends strongly on both of these quantities.  We therefore approximate the support of each distribution as the interval covering the observed range of values.  This is an underestimate of the true width, but allows for a consistent method of computing $\mathbb{D}$.

\subsubsection{Accurate quantification of extreme events}
Accurate quantification is indicated by models achieving low error on predictions from $\bx \in X$ such that $f(\bx)$ is rare.  We quantify this with the expected mean square error over the set of inputs corresponding to rare events.  For example, consider the mean square error restricted to the set of events with $p_y(y) < \epsilon$;
\begin{equation}
MSE_\epsilon = \bbE_{\bx}\left[\left. e(\bx)^2 \right| p_y(f(\bx)) \leq \epsilon \right],
\label{eq:MSE_eps}
\end{equation}
for values of $\epsilon > 0$.  Models that accurately quantify rare events should have low $MSE_\epsilon$ for $\epsilon \ll 1$. 
A more informative metric would condition on $p_y(f(\bx))=\epsilon$, but computation of such a quantity requires a parametric model or smoother. 
We therefore do not include it in this work.

\subsubsection{Accurate prediction of extreme events}
Models trained for regression may be used along with some threshold value to function as classifiers.  For a variety of extreme event rates, we track two metrics of classifier accuracy.  Following \cite{guth2019machine}, we use the extreme event rate dependent area under the precision recall curve given by,
\begin{equation}
\alpha (\omega; y, \hat{y}) = \int_\bbR s(1_{y \geq a}, 1_{\hat{y} \geq b}) \left| \frac{\partial}{\partial b} r(1_{y \geq a}, 1_{\hat{y} \geq b})\right| \, db
\label{eq:AUC}
\end{equation}
where $s$ and $r$ are the precision and recall and $a = F_y^{-1}(1-\omega)$ where $F_y^{-1}$ is the quantile function for $y$.  This quantity has in fact been used to train models with lower dimensional parameter spaces.  
However, the use of distinct thresholds for $y$ and $\hat{y}$ may be undesirable. Note for example that for any strictly increasing function $g$ we have $\alpha(\omega; y, \hat{y}) = \alpha(\omega; y, g(\hat{y}))$. 
We therefore also consider a metric that uses the same threshold for both $y$ and $\hat{y}$.  The extreme event rate dependent $F_1$ score is given by,
\begin{equation}
F_1 (\omega; y, \hat{y}) = F_1(1_{y \geq a}, 1_{\hat{y} \geq a})\hspace{5 mm} \text{where} \hspace{5mm} a = F_y^{-1}(1-\omega)
\label{eq:F1}
\end{equation}
where the $F_1$-score is given by the harmonic mean of precision and recall.

In many practical settings, the consequences of false positive or negative predictions will be disproportionate. 
In these cases, the metrics given by Eq. \eqref{eq:AUC} and \eqref{eq:F1} will not reflect the utility of the learned model and more setting specific metrics should be considered. 
\section{Results} \label{sec:results}

In this section we present the results of using each of the loss functions discussed in Sec. \ref{sec:methods} to two challenging supervised learning problems resulting from fluid dynamic systems with extreme events; the Kolmogorov flow at Reynolds number 40, and the flow around a square cylinder at Reynolds number 5000.  Numerical simulations of the incompressible Navier-Stokes equations in each case were performed using the spectral element method implemented in Nek5000 \cite{patera1984spectral, nek5000-web-page}, an open source Fortran code for incompressible fluids. 
Details on numerical simulation and data preparation are given in Appendix A.
In both cases, target data $y$ is centered and normalized to have zero mean and unit variance.

In each case we use LSTM networks implemented in tensorflow \cite{tensorflow2015-whitepaper} and trained using Adam \cite{kingma2014adam} with early stopping to avoid over-fitting. Datasets are split into training $(50\%)$, validation $(10\%)$ and testing $(40\%)$.  Each partition if formed from a contiguous set of samples so that phenomena observed in each are distinct.  Further details on training, as well as network structure are given in Appendix B.  We report the metrics outlined in Sec. \ref{subsec:error_measures} evaluated on the portion of the data reserved for testing.  

For each result we present data from twenty randomly initialized and trained networks.  Plots show mean value of across all trials as a solid line and shade region between $10^{th}$ and $90^{th}$ percentiles, thus excluding the two highest and two lowest values.

\subsection{Kolmogorov Flow}\label{subsec:kolmorogrov}

\begin{figure}
\centering
\includegraphics[width=\textwidth]{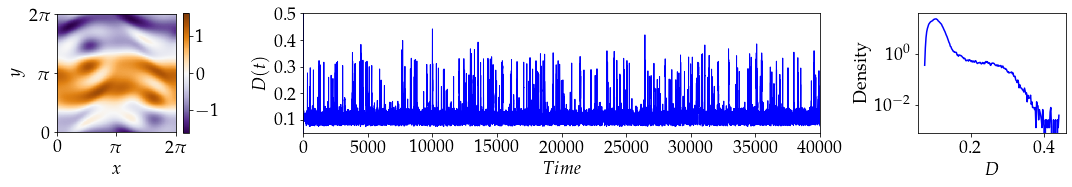}
\caption{Kolmogorov flow.  Left: snapshot of x-velocity.  Center: Time series for energy dissipation.  Right: Empirical density function of energy dissipation for $t\in [0,40,000]$.}
\label{fig:kol_u_D}
\end{figure}

We first consider two dimensional Kolmogorov flow at $Re=40$.  Dynamics follow the incompressible Navier-Stokes equations with a sinusoidal forcing term and periodic boundaries.  Specifically,
\begin{equation}
    \dfrac{\partial \bu}{\partial t} + \bu \cdot \nabla \bu = - \nabla p + \dfrac{1}{Re}\nabla^2 \bu + \sin (k_f y)\be_1 , \hspace{ 10 mm} \nabla \cdot \bu = 0,
    \label{eq:kol_flow}
\end{equation}
where $\be_1 = (1,0)$, $k_f=4$, and where we have boundary conditions $\bu(2\pi, y, t) = \bu(0,y,t)$ and $\bu(x, 2\pi, t) = \bu(x,0,t)$.
We are interested in this flow due to the behavior exhibited by the energy dissipation rate, given by a re-scaling of the enstrophy,
\begin{equation}
    D(t) = \frac{\nu}{|\Omega|} \int_\Omega |\nabla \bu(\bz, t)|^2 \, d\bz ,
    \label{eq:kol_energy_diss}
\end{equation}
where $\nu = Re^{-1}$ is the viscosity and $\Omega = [0, 2\pi)^2$ is the computational domain.
Solutions to Eq. \eqref{eq:kol_flow} with the prescribed boundary conditions are known to exhibit large bursts in energy input and dissipation rate resulting from intermittent alignment of the velocity field with the external forcing \cite{farazmand2016adjoint}.  
An single snapshot of the velocity field in the x-direction, as well as the time series for the energy dissipation rate and its density are shown in Fig. \ref{fig:kol_u_D}.

The intermittent bursts apparent in Fig. \ref{fig:kol_u_D} have been the subject of several previous works seeking to predict their occurrence in advance \cite{farazmand2016adjoint, guth2019machine, asch2021model}. 
In particular, Farazmand and Sapsis \cite{farazmand2017variational} showed that the intermittent behavior could be explained through triadic interactions between Fourier modes.  Here we consider the same problem, but seek to predict bursts in the energy dissipation using neural networks.  The target ($y$) quantity is the energy dissipation rate $D(t)$, normalized to have zero mean and unit variance. 
We use the time varying magnitudes of the three Fourier modes identified in \cite{farazmand2017variational} as inputs; $\bx(t) = (a_{0,k_f}(t),a_{1, 0}(t), a_{1, k_f}(t))$.  In order to ensure accurate statistical descriptions of errors, a simulation of the Kolmogorov flow run for $4\cdot 10^4$ time units.

\begin{figure}[t!]
\centering
\includegraphics[width=\textwidth]{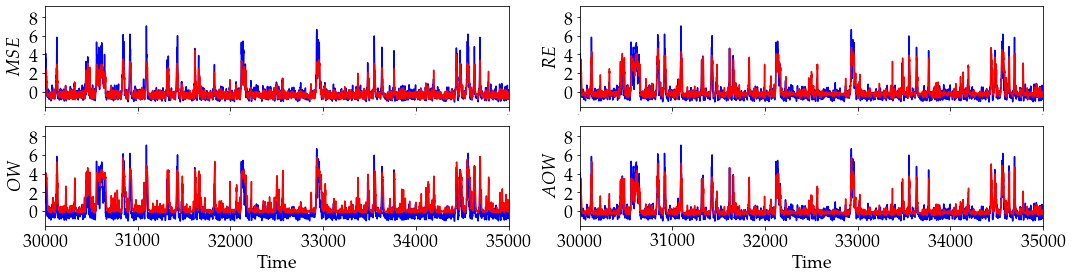}
\caption{Mean predictions for subset of test data for the Kolmogorov dataset at lead time $\tau=6$.  Blue time series is true $D(t)$ and red is mean prediction across 20 trained neural networks.}
\label{fig:kol_time_series}
\end{figure}

\begin{figure}
\centering
\includegraphics[width=\textwidth]{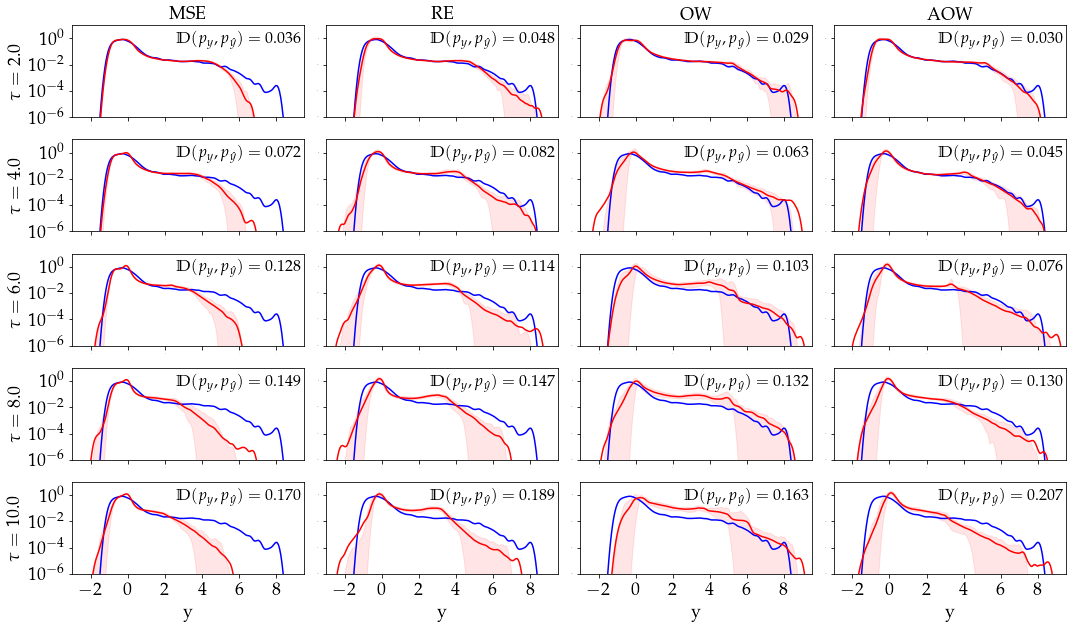}
\caption{Density for the Kolmogorov flow energy dissipation using true data (blue) and 20 realizations of neural networks trained with each loss function (red).  Reported $\bbD$ is mean of Eq. \eqref{eq:log_diff_pdf}.}
\label{fig:kol_density}
\end{figure}

We train neural network predictors of $D(t+\tau)$ from $\{\bx(s) : s\leq t\}$ for several lead times $\tau$.  LSTM networks \cite{hochreiter1997long} are used to allow for historical data to assist in prediction. 
Further details on the neural network structure and training procedure may be found in Appendix B. 
For each network configuration and lead time, we train twenty networks from randomly initialized weights.
A subset of the test data results is shown for a lead time of $\tau=6$ in Fig. \ref{fig:kol_time_series}. 
The true value of $D(t)$ is shown in blue and the mean prediction across twenty networks for each loss function is shown in red.
A cursory inspection reveals that $L_{MSE}$ appears to underestimate the large fluctuations, though it does appear to predict their locations.  Each of the three other loss functions appear to quantify the peaks more accurately, with $L_{OW}$ having the most pronounced false positives. 

Estimates of the densities $p_y$ using test set values of $y_i=f(\bx_i)$ and predictions $\hat{y}_i=\hat{f}(\bx_i)$ with $\hat{f}$ trained using each of the loss functions considered in this work are shown in Fig. \ref{fig:kol_density}.
The normalized difference between logs metric $\mathbb{D}$ is also shown alongside each loss function and lead time.
At each lead time, the induced density using models trained with $L_{MSE}$ has the highest fidelity in the core of the pdf, but these distributions dramatically underestimate the density in the high dissipation tail region.
At a low lead time of $\tau = 2$, densities from models trained using $L_{RE}$, $L_{OW}$, and $L_{AOW}$ perform approximately equally.
For higher $\tau$, $L_{RE}$ and $L_{AOW}$ outperform $L_{OW}$, with the latter tending to overestimate the density positive outliers. 
Excepting $\tau=10.0$, the minimal value of Eq. \eqref{eq:log_diff_pdf} is obtained by $L_{AOW}$. 
At $\tau=10$ the best numerical result is obtained using $L_{MSE}$. 
However, the estimated density function reveals significant underestimation of tail events. 

\begin{figure}
\centering
\includegraphics[width=\textwidth]{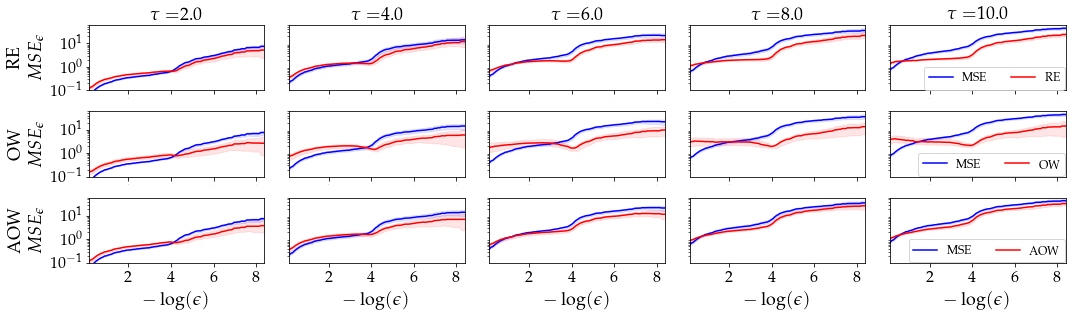}
\caption{Comparison of $MSE_\epsilon$ for models trained using $L_{MSE}$ (blue) and extreme event specific loss functions (red) for the Kolmogorov dataset.  Note that high $-\log(\epsilon)$ corresponds to rare events.}
\label{fig:kol_mse_eps_small}
\end{figure}

\begin{figure}
\centering
\includegraphics[width=\textwidth]{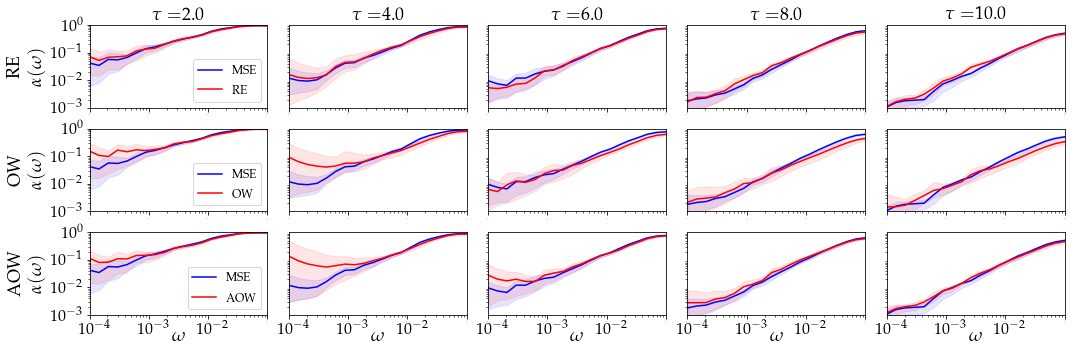}
\caption{Comparison of $\alpha (\omega)$ for models trained using $L_{MSE}$ (blue) and extreme event specific loss functions (red) for the Kolmogorov dataset.}
\label{fig:kol_alpha_small}
\end{figure}

The rare event error given by Eq. \eqref{eq:MSE_eps} is shown in Fig. \ref{fig:kol_mse_eps_small}.  Each subplot includes error statistics for twenty networks trained with $L_{MSE}$ in blue and networks trained using each of the specialized loss functions in red.
In all cases, $L_{MSE}$ has lower error when $\epsilon$ is small, which is unsurprising since this is the objective $L_{MSE}$ minimizes.
The difference is more pronounced when compared to $L_{OW}$ or when $\tau$ is small.
For rare events, each of the specialized loss functions has lower error.

\begin{figure}
\centering
\includegraphics[width=\textwidth]{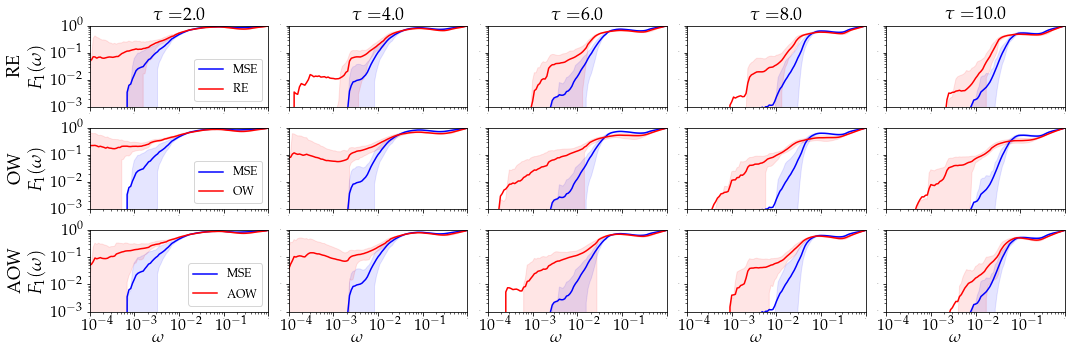}
\caption{Comparison of $F_1 (\omega)$ for models trained using $L_{MSE}$ (blue) and extreme event specific loss functions (red) for the Kolmogorov dataset.  Lines indicates mean and shaded region is bounded by 10th and 90th percentiles.}
\label{fig:kol_F1}
\end{figure}

The performance of the trained networks as classifiers for rare events is shown in Figs. \ref{fig:kol_alpha_small} and \ref{fig:kol_F1}.
Fig. \ref{fig:kol_alpha_small} shows that area under the precision recall curve as a function of extreme event rate.
At low lead times and extreme event rates, $L_{OW}$ and $L_{AOW}$ show some improvement over MSE, but all methods are approximately equivalent in other cases. 
Note however, that $\alpha (\omega)$ allows for alternative thresholds to be applied to $y$ and $\hat{y}$.
Fig. \ref{fig:kol_F1} shows very clearly that across all cases the extreme event specific loss functions yield models that are more accurately able to distinguish extreme events when the same threshold is applied.

\subsection{Flow around a square cylinder}\label{subsec:square}

\begin{figure}
\centering
\includegraphics[width=\textwidth]{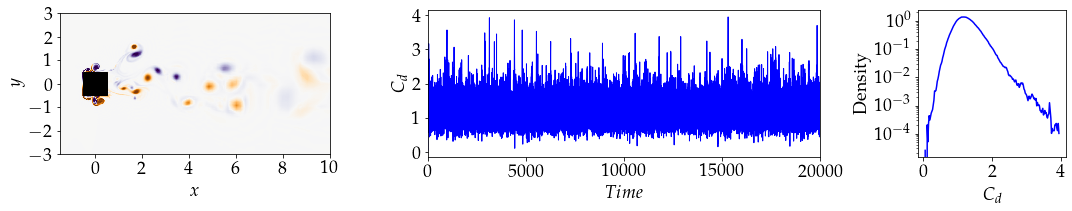}
\caption{Flow around a square cylinder at $Re=5000$.  Left: Snapshot of vorticity close to cylinder.  Center: Time series for drag coefficient.  Right: Empirical density function of drag coefficient.}
\label{fig:square_vort_drag}
\end{figure}

Our second example considers the flow around a square cylinder at Reynolds number $Re=5000$.  The square has unit side-length and is positioned in a stream having unit inlet velocity and viscosity $\nu = 2\cdot 10^{-4}$. 
The flow is characterized by extremely chaotic vortex shedding in the wake of the cylinder, shown in the left panel of Fig. \ref{fig:square_vort_drag} and large deviations in the forcing applied to the cylinder by the fluid.  The drag coefficient is a non-dimensional quantity given by,
\begin{equation}
    C_d(t) = \dfrac{2}{\rho u_\infty^2 C}\oint_{\partial S} \left(\boldsymbol{\tau}(t) - p(t)\bn \right) \cdot \be_x \, ds
    \label{eq:drag}
\end{equation}
where $\boldsymbol{\tau}$, $p$, and $\bn$ are the skin shear stress, pressure, and normal vector, and $\partial S$ is the boundary of the square cylinder.  $\rho$, $u_\infty$, and $C$ all have numerical value of 1 and are included only for their dimension.  The time series of $C_d(t)$ for a simulation of length $T=2\cdot 10^4$ and empirical density function are shown in the center and right panels of Fig. \ref{fig:square_vort_drag}.

We consider the problem of predicting $C_d(t+\tau)$, centered and normalized to unit variance, from 40 evenly spaced measurements of pressure on the surface of the cylinder, $P(t) \in \bbR^{40}$.
The contributions of $\boldsymbol{\tau}$ and $p(t)$ to $C_d$ are called the skin friction drag and pressure drag, respectively.
In this example, pressure drag is several orders of magnitude larger than skin friction drag. 
Therefore, assuming a sufficiently dense distribution of pressure measurements prediction of $C_d$ at zero lead time is proportional to the difference of average pressure on the front and back of the square. 
However, for $\tau > 0$ the problem rapidly becomes challenging.

\begin{figure}
\centering
\includegraphics[width=\textwidth]{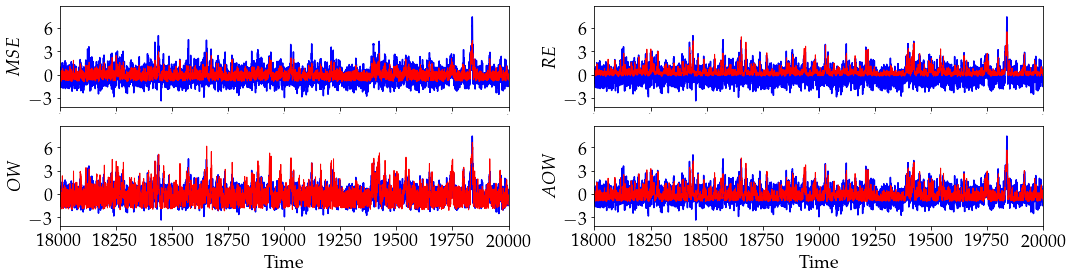}
\caption{True centered and normalized drag coefficient (blue) and mean time series predictions (red) for test set data from each neural network for the drag on a square cylinder.  Results are for a lead time of $\tau=1$.}
\label{fig:square_time_series}
\end{figure}

We test of of the loss functions proposed in Sec. \ref{sec:methods} using an LSTM network described in further detail in Appendix B. 
Time series of the true drag coefficient as well as the mean prediction from twenty neural networks trained using each of the loss functions are shown in Fig. \ref{fig:square_time_series} for a lead time of $\tau=1$. 
It is immediately clear that predictions made by networks trained with different loss functions exhibit substantially different behavior.
Consistent with the Kolmogorov flow data, $L_{MSE}$ substantially underestimates fluctuations, while $L_{OW}$ has many false positives.

\begin{figure}
\centering
\includegraphics[width=\textwidth]{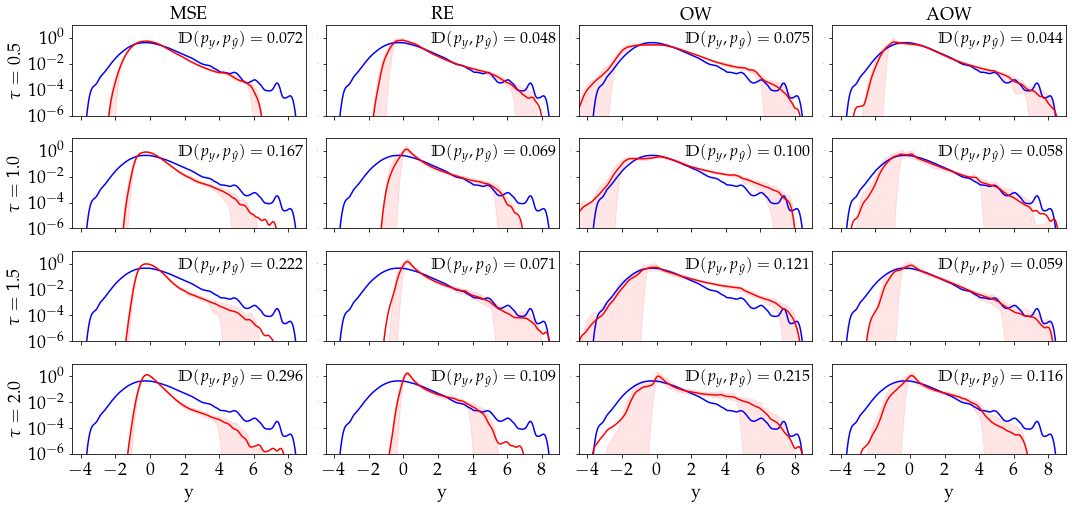}
\caption{Density functions for the drag on a square cylinder using true data (blue) and 20 realizations of neural networks trained with each loss function (red).}
\label{fig:square_density}
\end{figure}

Estimates of the density of $y$ and of $\hat{y}$ formed from test set data for each loss function and several lead times are shown in Fig. \ref{fig:square_density}. 
As expected, networks trained with $L_{MSE}$ underestimate the likelihood of events in the tails of the distribution of $y$ while $L_{OW}$ overestimates these same events. 
$L_{RE}$ and $L_{AOW}$ generally both capture the correct tail behavior for extreme events skewing positive, with $L_{AOW}$ achieving lower $\mathbb{D}$. 
We note that while $L_{RE}$ performs poorly for quantifying $p_y$ when $y<0$, this may be a consequence of our choice of parameter $\lambda$, which weighed positive values ten times as much as negative.
In each case, $L_{RE}$ and $L_{AOW}$ notable outperform $L_{OW}$ which in turn outperforms $MSE$ with equards to mean $\bbD$.

\begin{figure}
\centering
\includegraphics[width=\textwidth]{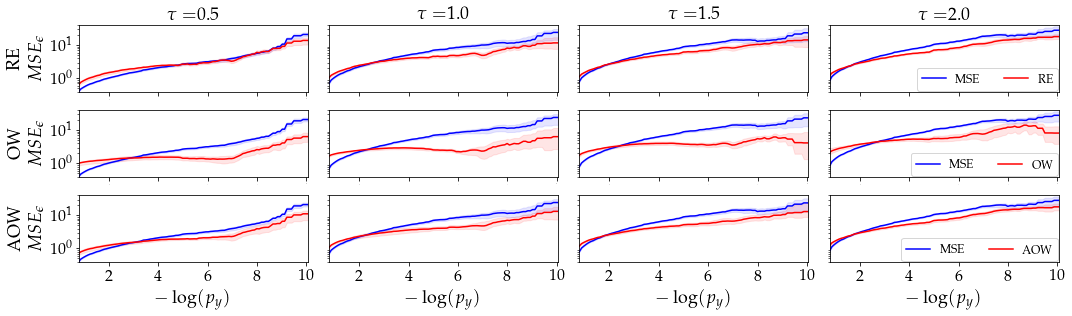}
\caption{Comparison of $MSE_\epsilon$ for models trained using $L_{MSE}$ (blue) and extreme event specific loss functions (red) for the square dataset.  Note, high $-\log(\epsilon)$ corresponds to rare events.}
\label{fig:square_mse_eps}
\end{figure}

Plots of $MSE_\epsilon$ are shown in Fig. \ref{fig:square_mse_eps}. 
As expected loss functions tailored to extreme events are better able to quantify events occurring with a low probability (high $-\log(\epsilon)$), while underperforming $L_{MSE}$ for more frequent events.
This is consistent with expectations and supports the use of specialized loss functions for quantifying rare events.

\begin{figure}
\centering
\includegraphics[width=\textwidth]{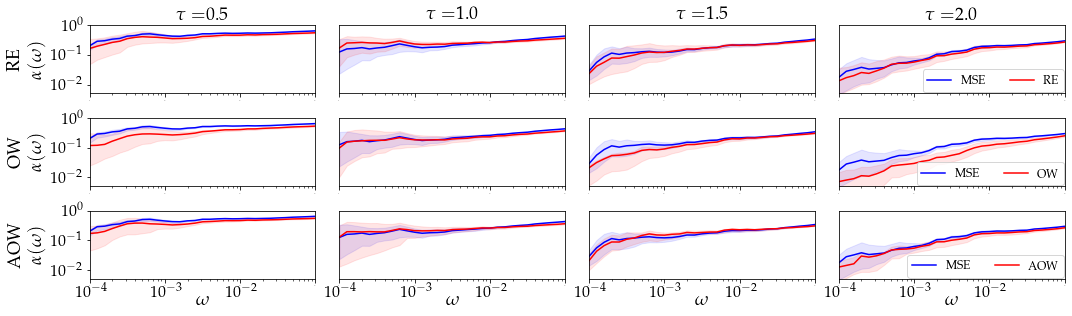}
\caption{Comparison of $\alpha (\omega)$ for models trained using $L_{MSE}$ (blue) and extreme event specific loss functions (red) for the square dataset.}
\label{fig:square_alpha}
\end{figure}

\begin{figure}
\centering
\includegraphics[width=\textwidth]{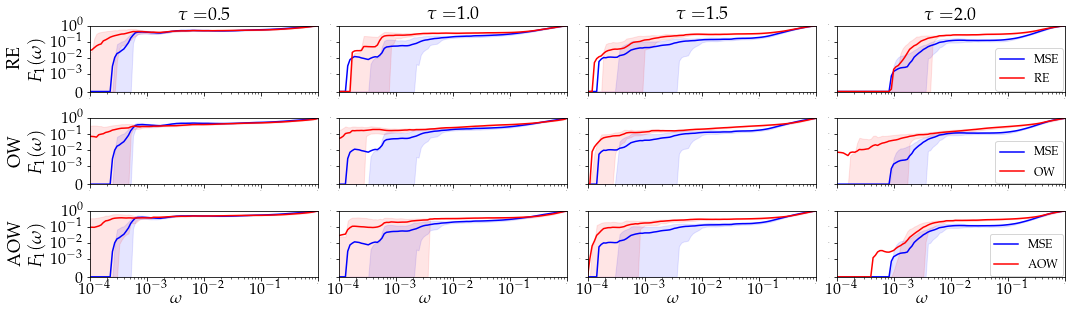}
\caption{Comparison of $F_1 (\omega)$ for models trained using $L_{MSE}$ (blue) and extreme event specific loss functions (red) for the square dataset.}
\label{fig:square_F1_same}
\end{figure}

Classification metrics for the square data are shown in Figs. \ref{fig:square_alpha} and \ref{fig:square_F1_same}. 
Figure \ref{fig:square_alpha} provides evidence that challenges the utility of the extreme event specific loss functions. 
At a lead time of $\tau=1$, $L_{RE}$ slightly outperforms $L_{MSE}$ in the low extreme event rate region. 
However, $L_{MSE}$ has the highest $\alpha (\omega)$ in all other cases. 
This suggests that while networks trained $L_{MSE}$ do a poor job at quantifying extreme events for this example, they are in fact able to do a better job than others at separating extreme from quiescent events given an appropriate pair of thresholds. 
However, this is not true when the same threshold is applied to each of $y$ and $\hat{y}$. 
Figure \ref{fig:square_F1_same} shows that each of the extreme event specific loss functions outperforms $L_{MSE}$ when the same threshold is used.
\section{Discussion} \label{sec:discussion}

In this work we have developed and evaluated several candidate loss functions for use in regression problems seeking to accurately quantify and predict extreme events.
We have taken care to describe the motivation for each of the loss functions as an approximation of a continuous functional, allowing for greater intuition and providing a framework upon which further improvements may be made.
The test cases used to evaluate the loss functions include the Kolmogorov flow, which has been studied extensively as a canonical example of a turbulent fluid flow exhibiting extreme events and the flow around a square cylinder at $Re=5000$. 
This latter example has, to the best of our knowledge, not been studied in the context of extreme events. 
In each case, the proposed loss functions yield models with significantly improved error in the tails of the output, more accurate estimates of the tail likelihoods, and improved classification when an equal threshold is applied to prediction and ground truth. 

We have also provided evidence, however, that prediction and quantification of extreme events are distinct problems.
Figure \ref{fig:square_alpha} makes it clear that improved performance in extreme event quantification does not necessarily mean greater performance in classification, as measured by the area under the precision recall curve. 
However, this result requires distinct thresholds for what is considered an extreme event under $f$ and $\hat{f}$. 
Thus, the area under the precision recall curve interprets $\hat{f}$ not as an interpolation of $f$ but as a distinct metric used for classification.

The ``best'' loss function for a particular task will depend on the problem, dataset, and goals. 
Of those studied in this work, $L_{MSE}$ and $L_{RE}$ have the advantage of not requiring an estimate of $p_y$. 
The extent to which this proves troublesome will depend on the dimension of $y$, sample size, and on the distribution $p_y$.
Tail events are most accurately quantified by $L_{OW}$, though at the cost of poor performance on events in the core of $p_y$.  This is partially mitigated by $L_{AOW}$. 
Networks used in any application setting should be selected via cross-validation using a problem-appropriate metric.

An obvious challenge in the use of loss functions which include $p_y$ is that the number of samples required for density approximation grows geometrically in the dimension of $y$. 
In this work we have focused on the case where $y$ is scalar valued and thus avoided the issue. 
Problems in higher dimensions may require approximating density through some observable or via distance to neighbors along a low dimensional manifold via technique like diffusion maps \cite{coifman2006diffusion}, if such a manifold exists. 
The relative entropy loss does not require approximation of $p_y$, but does equate importance of a particular sample with its exponentiated magnitude. 
Thus, it may not be an ideal choice when the rare events of interest are not substantially different in magnitude from the core of the distribution.
Further investigation of such problems would be an interesting research direction, but we consider it to be outside the scope of this work. 

The proposed extreme event loss functions may be a poor choice for certain classes of prediction problems, even if the data has extreme events.
Consider for example the task of learning a dynamic model for a system with extreme events using short time series or velocity data as in \cite{doan2021short,wan2018data}. 
In this scenario, errors in the prediction phase are compounded and the higher error rate in the core of the distribution will render long term forecasts less accurate. 
This is in contrast to models trained with $L_{MSE}$ where error accumulation will likely be more focused on rare events.

There exists a wide spectrum of functional forms used in machine learning that allows researchers to select or construct models they deem fit for a particular task. 
Less attention is paid to the choice of functional indicating the performance of those models, though some works have considered this question and proved its importance \cite{qi2020using, doan2021short}. 
This may be due to the wide effectiveness of the often used mean square error, and its clear motivation as a miximum likelihood estimate given Gaussian error. 
However, as we have shown, performance on certain tasks is significantly improved with tailored loss functions.
The present manuscript extends this important research direction. \\

\noindent \textbf{Funding:}\\

This material is based upon work supported by the National Science Foundation (1902972),
the Office of Navy Research (N00014-21-1-2357),
and the Airforce Office of Scientific Research (MURI FA9550-21-1-0058).
Simulation of the flow around a square cylinder was run using the Extreme Science and Engineering Discovery Environment (XSEDE) \cite{towns2014xsede} allocation TG-MTH210003.  XSEDE is supported by National Science Foundation under Grant No. ACI-1548562.

\section*{Appendix A: Details on datasets}\label{app:data}

Numerical simulations of the Kolmogorov flow and flow around a square cylinder were performed using Nek5000 \cite{nek5000-web-page}, an open source Fortran based spectral element code for incompressible flow \cite{patera1984spectral}.  Time integration was caried out using a second order semi-implicit scheme described in \cite{fischer2003implementation}.  Simulations were stabilised with a small degree of filtering as described in \cite{fischer2001filter}.  Mesh was generated using gmsh \cite{geuzaine2009gmsh}.  The Kolmorogorov flow dataset uses 144 elements of order 7 and the flow around a square cylinder uses 1728 elements of order 7.  The domain for the square cylinder flow extended from $x=-12$ to $x=30$ and $|y|\leq 12$ with the square having sidelength $1$ centered at the origin.  Geometry and case files for rerunning simulations as well as files with numerical values of output data used in this work are available on GitHub at \url{https://github.com/snagcliffs/EE_loss}.

The mean and standard deviation of the energy dissipation for the Kolmogorov flow dataset used in this work are $0.116065$ and $0.037559$, which are respectively within $0.64\%$ and $2.1\%$ of those reported in \cite{farazmand2017variational}.
The grid for the flow around a square cylinder was validated via comparison to a short simulation using a finer mesh.  The mean drag coefficient on the interval $t \in [200,2000]$ using 1728 spectral elements was found to be approximately $1.34\%$ different from that using 4480 elements, which was deemed sufficiently resolved for the purpose of this work.

\section*{Appendix B: Neural networks details}\label{app:nn}

Neural networks used in this manuscript were implemented in Python using the tensorflow library \cite{tensorflow2015-whitepaper} as well as the Numba library \cite{lam2015numba}. 
Gaussian process models for $p_y$ were initially trained using GPy \cite{gpy2014}, a Python library for Gaussian processes. 
Learned parameters were subsequently used to build non-trainable Gaussian process models in tensorflow, allowing for their use with neural network optimization tools.
Source code for neural networks and scripts used to train examples used in this work are available on GitHub at \url{https://github.com/snagcliffs/EE_loss}.

Networks were built using a combination of dense layers and long-short-term-memory (LSTM) layers \cite{hochreiter1997long}. 
Networks for the Kolmogrov flow had 3 dense layers of size (4,8,16), followed by an LSTM layer with (32) units, followed by dense layers of size (16,8,4).
Networks for the square cylinder had 3 dense layers of size (4,8,16), followed by an LSTM layer with (16) units, followed by dense layers of size (16,8,4).
In each case the swish activation function was used \cite{ramachandran2017searching}.

Neural networks are widely known to be prone to over fitting. 
To mitigate this, we used early stopping \cite{morgan1989generalization} and noise injection in the training data. 
Data was separated into disjoint temporally contiguous sets for training, validation, and testing using a (50/10/40\%) split and training was stopped when validation set loss failed to yield a new minimum for 5 consecutive epochs.
Noise injection has been shown to improve generalization performance \cite{an1996effects, asch2021model}, which may be due in part to its relation to Tikhonov regularization \cite{bishop1995training}. 
During the training procedure, we sampled random noise from a Gaussian distribution having standard deviation equal to 10\% of the data and added it to both the neural network input.

\small
\begin{spacing}{.5}
\bibliographystyle{plain}
\bibliography{refs}

\begin{thebibliography}{10}

\bibitem{tensorflow2015-whitepaper}
Mart\'{\i}n Abadi and et. al.
\newblock {TensorFlow}: Large-scale machine learning on heterogeneous systems,
  2015.
\newblock Software available from tensorflow.org.

\bibitem{an1996effects}
Guozhong An.
\newblock The effects of adding noise during backpropagation training on a
  generalization performance.
\newblock {\em Neural computation}, 8(3):643--674, 1996.

\bibitem{asch2021model}
Anna Asch, Ethan Brady, Hugo Gallardo, John Hood, Bryan Chu, and Mohammad
  Farazmand.
\newblock Model-assisted deep learning of rare extreme events from partial
  observations.
\newblock {\em arXiv preprint arXiv:2111.04857}, 2021.

\bibitem{bishop1995training}
Chris~M Bishop.
\newblock Training with noise is equivalent to tikhonov regularization.
\newblock {\em Neural computation}, 7(1):108--116, 1995.

\bibitem{blanchard2021bayesian}
Antoine Blanchard and Themistoklis Sapsis.
\newblock Bayesian optimization with output-weighted optimal sampling.
\newblock {\em Journal of Computational Physics}, 425:109901, 2021.

\bibitem{blonigan2019extreme}
Patrick~J Blonigan, Mohammad Farazmand, and Themistoklis~P Sapsis.
\newblock Are extreme dissipation events predictable in turbulent fluid flows?
\newblock {\em Physical Review Fluids}, 4(4):044606, 2019.

\bibitem{brenner2019perspective}
MP~Brenner, JD~Eldredge, and JB~Freund.
\newblock Perspective on machine learning for advancing fluid mechanics.
\newblock {\em Physical Review Fluids}, 4(10):100501, 2019.

\bibitem{brenowitz2018prognostic}
Noah~D Brenowitz and Christopher~S Bretherton.
\newblock Prognostic validation of a neural network unified physics
  parameterization.
\newblock {\em Geophysical Research Letters}, 45(12):6289--6298, 2018.

\bibitem{brunton2020machine}
Steven~L Brunton, Bernd~R Noack, and Petros Koumoutsakos.
\newblock Machine learning for fluid mechanics.
\newblock {\em Annual Review of Fluid Mechanics}, 52:477--508, 2020.

\bibitem{champion2019data}
Kathleen Champion, Bethany Lusch, J~Nathan Kutz, and Steven~L Brunton.
\newblock Data-driven discovery of coordinates and governing equations.
\newblock {\em Proceedings of the National Academy of Sciences},
  116(45):22445--22451, 2019.

\bibitem{chang1997conditioning}
Joseph~T Chang and David Pollard.
\newblock Conditioning as disintegration.
\newblock {\em Statistica Neerlandica}, 51(3):287--317, 1997.

\bibitem{coifman2006diffusion}
Ronald~R Coifman and St{\'e}phane Lafon.
\newblock Diffusion maps.
\newblock {\em Applied and computational harmonic analysis}, 21(1):5--30, 2006.

\bibitem{doan2021short}
NAK Doan, W~Polifke, and L~Magri.
\newblock Short-and long-term predictions of chaotic flows and extreme events:
  a physics-constrained reservoir computing approach.
\newblock {\em Proceedings of the Royal Society A}, 477(2253):20210135, 2021.

\bibitem{duraisamy2019turbulence}
Karthik Duraisamy, Gianluca Iaccarino, and Heng Xiao.
\newblock Turbulence modeling in the age of data.
\newblock {\em Annual Review of Fluid Mechanics}, 51:357--377, 2019.

\bibitem{dysthe2008oceanic}
Kristian Dysthe, Harald~E Krogstad, and Peter M{\"u}ller.
\newblock Oceanic rogue waves.
\newblock {\em Annu. Rev. Fluid Mech.}, 40:287--310, 2008.

\bibitem{easterling2000observed}
David~R Easterling, J~L Evans, P~Ya Groisman, Thomas~R Karl, Kenneth~E Kunkel,
  and P~Ambenje.
\newblock Observed variability and trends in extreme climate events: a brief
  review.
\newblock {\em Bulletin of the American Meteorological Society},
  81(3):417--426, 2000.

\bibitem{efron1981nonparametric}
Bradley Efron.
\newblock Nonparametric standard errors and confidence intervals.
\newblock {\em Canadian Journal of Statistics}, 9(2):139--158, 1981.

\bibitem{farazmand2016adjoint}
Mohammad Farazmand.
\newblock An adjoint-based approach for finding invariant solutions of
  navier--stokes equations.
\newblock {\em Journal of Fluid Mechanics}, 795:278--312, 2016.

\bibitem{farazmand2017variational}
Mohammad Farazmand and Themistoklis~P Sapsis.
\newblock A variational approach to probing extreme events in turbulent
  dynamical systems.
\newblock {\em Science advances}, 3(9):e1701533, 2017.

\bibitem{farazmand2019extreme}
Mohammad Farazmand and Themistoklis~P Sapsis.
\newblock Extreme events: Mechanisms and prediction.
\newblock {\em Applied Mechanics Reviews}, 71(5), 2019.

\bibitem{fischer2001filter}
Paul Fischer and Julia Mullen.
\newblock Filter-based stabilization of spectral element methods.
\newblock {\em Comptes Rendus de l'Acad{\'e}mie des Sciences-Series
  I-Mathematics}, 332(3):265--270, 2001.

\bibitem{fischer2003implementation}
PF~Fischer.
\newblock Implementation considerations for the oifs/characteristics approach
  to convection problems.
\newblock {\em Argonne National Laboratory}, 2003.

\bibitem{geuzaine2009gmsh}
Christophe Geuzaine and Jean-Fran{\c{c}}ois Remacle.
\newblock Gmsh: A 3-d finite element mesh generator with built-in pre-and
  post-processing facilities.
\newblock {\em International journal for numerical methods in engineering},
  79(11):1309--1331, 2009.

\bibitem{goodfellow2016deep}
Ian Goodfellow, Yoshua Bengio, and Aaron Courville.
\newblock {\em Deep learning}.
\newblock MIT press, 2016.

\bibitem{gpy2014}
{GPy}.
\newblock {GPy}: A gaussian process framework in python.
\newblock \url{http://github.com/SheffieldML/GPy}, since 2012.

\bibitem{gupta2020neural}
Abhinav Gupta and Pierre~FJ Lermusiaux.
\newblock Neural closure models for dynamical systems.
\newblock {\em Proceedings of the Royal Society A}, 477(2252):20201004, 2021.

\bibitem{guth2019machine}
Stephen Guth and Themistoklis~P Sapsis.
\newblock Machine learning predictors of extreme events occurring in complex
  dynamical systems.
\newblock {\em Entropy}, 21(10):925, 2019.

\bibitem{he2009learning}
Haibo He and Edwardo~A Garcia.
\newblock Learning from imbalanced data.
\newblock {\em IEEE Transactions on knowledge and data engineering},
  21(9):1263--1284, 2009.

\bibitem{hochreiter1997long}
Sepp Hochreiter and J{\"u}rgen Schmidhuber.
\newblock Long short-term memory.
\newblock {\em Neural computation}, 9(8):1735--1780, 1997.

\bibitem{irrgang2021towards}
Christopher Irrgang, Niklas Boers, Maike Sonnewald, Elizabeth~A Barnes,
  Christopher Kadow, Joanna Staneva, and Jan Saynisch-Wagner.
\newblock Towards neural earth system modelling by integrating artificial
  intelligence in earth system science.
\newblock {\em Nature Machine Intelligence}, 3(8):667--674, 2021.

\bibitem{kingma2014adam}
Diederik~P Kingma and Jimmy Ba.
\newblock Adam: A method for stochastic optimization.
\newblock {\em arXiv preprint arXiv:1412.6980}, 2014.

\bibitem{lam2015numba}
Siu~Kwan Lam, Antoine Pitrou, and Stanley Seibert.
\newblock Numba: A llvm-based python jit compiler.
\newblock In {\em Proceedings of the Second Workshop on the LLVM Compiler
  Infrastructure in HPC}, pages 1--6, 2015.

\bibitem{longin1996asymptotic}
Fran{\c{c}}ois~M Longin.
\newblock The asymptotic distribution of extreme stock market returns.
\newblock {\em Journal of business}, pages 383--408, 1996.

\bibitem{lu2021learning}
Lu~Lu, Pengzhan Jin, Guofei Pang, Zhongqiang Zhang, and George~Em Karniadakis.
\newblock Learning nonlinear operators via deeponet based on the universal
  approximation theorem of operators.
\newblock {\em Nature Machine Intelligence}, 3(3):218--229, 2021.

\bibitem{lusch2018deep}
Bethany Lusch, J~Nathan Kutz, and Steven~L Brunton.
\newblock Deep learning for universal linear embeddings of nonlinear dynamics.
\newblock {\em Nature communications}, 9(1):1--10, 2018.

\bibitem{majda1997one}
AJ~Majda, DW~McLaughlin, and EG1431687 Tabak.
\newblock A one-dimensional model for dispersive wave turbulence.
\newblock {\em Journal of Nonlinear Science}, 7(1):9--44, 1997.

\bibitem{mchutchon2013differentiating}
Andrew McHutchon.
\newblock Differentiating gaussian processes.
\newblock {\em Cambridge (ed.)}, 2013.

\bibitem{milano2002neural}
Michele Milano and Petros Koumoutsakos.
\newblock Neural network modeling for near wall turbulent flow.
\newblock {\em Journal of Computational Physics}, 182(1):1--26, 2002.

\bibitem{mohamad2018sequential}
Mustafa~A Mohamad and Themistoklis~P Sapsis.
\newblock Sequential sampling strategy for extreme event statistics in
  nonlinear dynamical systems.
\newblock {\em Proceedings of the National Academy of Sciences},
  115(44):11138--11143, 2018.

\bibitem{morgan1989generalization}
Nelson Morgan and Herv{\'e} Bourlard.
\newblock Generalization and parameter estimation in feedforward nets: Some
  experiments.
\newblock {\em Advances in neural information processing systems}, 2:630--637,
  1989.

\bibitem{patera1984spectral}
Anthony~T Patera.
\newblock A spectral element method for fluid dynamics: laminar flow in a
  channel expansion.
\newblock {\em Journal of computational Physics}, 54(3):468--488, 1984.

\bibitem{nek5000-web-page}
James W.~Lottes Paul F.~Fischer and Stefan~G. Kerkemeier.
\newblock {nek5000} {W}eb page, 2008.
\newblock http://nek5000.mcs.anl.gov.

\bibitem{qi2020using}
Di~Qi and Andrew~J Majda.
\newblock Using machine learning to predict extreme events in complex systems.
\newblock {\em Proceedings of the National Academy of Sciences}, 117(1):52--59,
  2020.

\bibitem{qian2020lift}
Elizabeth Qian, Boris Kramer, Benjamin Peherstorfer, and Karen Willcox.
\newblock Lift \& learn: Physics-informed machine learning for large-scale
  nonlinear dynamical systems.
\newblock {\em Physica D: Nonlinear Phenomena}, 406:132401, 2020.

\bibitem{raissi2019physics}
Maziar Raissi, Paris Perdikaris, and George~E Karniadakis.
\newblock Physics-informed neural networks: A deep learning framework for
  solving forward and inverse problems involving nonlinear partial differential
  equations.
\newblock {\em Journal of Computational Physics}, 378:686--707, 2019.

\bibitem{ramachandran2017searching}
Prajit Ramachandran, Barret Zoph, and Quoc~V Le.
\newblock Searching for activation functions.
\newblock {\em arXiv preprint arXiv:1710.05941}, 2017.

\bibitem{rasmussen2003gaussian}
Carl~Edward Rasmussen.
\newblock Gaussian processes in machine learning.
\newblock In {\em Summer school on machine learning}, pages 63--71. Springer,
  2003.

\bibitem{rasp2018deep}
Stephan Rasp, Michael~S Pritchard, and Pierre Gentine.
\newblock Deep learning to represent subgrid processes in climate models.
\newblock {\em Proceedings of the National Academy of Sciences},
  115(39):9684--9689, 2018.

\bibitem{sapsis2020output}
Themistoklis~P Sapsis.
\newblock Output-weighted optimal sampling for bayesian regression and rare
  event statistics using few samples.
\newblock {\em Proceedings of the Royal Society A}, 476(2234):20190834, 2020.

\bibitem{sapsis2021statistics}
Themistoklis~P Sapsis.
\newblock Statistics of extreme events in fluid flows and waves.
\newblock {\em Annual Review of Fluid Mechanics}, 53:85--111, 2021.

\bibitem{siegmund1976importance}
David Siegmund.
\newblock Importance sampling in the monte carlo study of sequential tests.
\newblock {\em The Annals of Statistics}, pages 673--684, 1976.

\bibitem{towns2014xsede}
John Towns, Timothy Cockerill, Maytal Dahan, Ian Foster, Kelly Gaither, Andrew
  Grimshaw, Victor Hazlewood, Scott Lathrop, Dave Lifka, Gregory~D Peterson,
  et~al.
\newblock Xsede: accelerating scientific discovery.
\newblock {\em Computing in science \& engineering}, 16(5):62--74, 2014.

\bibitem{wan2018data}
Zhong~Yi Wan, Pantelis Vlachas, Petros Koumoutsakos, and Themistoklis Sapsis.
\newblock Data-assisted reduced-order modeling of extreme events in complex
  dynamical systems.
\newblock {\em PloS one}, 13(5):e0197704, 2018.

\bibitem{wasserman2006all}
Larry Wasserman.
\newblock {\em All of nonparametric statistics}.
\newblock Springer Science \& Business Media, 2006.

\bibitem{yeung2015extreme}
PK~Yeung, XM~Zhai, and Katepalli~R Sreenivasan.
\newblock Extreme events in computational turbulence.
\newblock {\em Proceedings of the National Academy of Sciences},
  112(41):12633--12638, 2015.

\end{thebibliography}
\end{spacing}

\end{document}